\documentclass{article}

\usepackage[preprint, nonatbib]{neurips_2026}
\usepackage[utf8]{inputenc}
\usepackage[T1]{fontenc}
\usepackage{amsmath,amssymb,amsfonts}
\usepackage{nicefrac}
\usepackage{booktabs}
\usepackage{graphicx}
\graphicspath{{images/}}
\usepackage{hyperref}
\usepackage{url}
\usepackage{microtype}
\usepackage{xcolor}
\usepackage{enumitem}
\usepackage{array}
\usepackage{multirow}
\usepackage{tabularx}

\hypersetup{
  colorlinks=true,
  linkcolor=blue!60!black,
  citecolor=blue!60!black,
  urlcolor=blue!60!black,
}


\title{Trust Between AI Agents:\\Measuring Formation, Breakage, and Recovery,\\with Implications for Governing Multi-Agent Systems}

\author{
  Yujiao Chen \\
  Massachusetts Institute of Technology \\
  Cambridge, MA 02139 \\
  \texttt{yujiaoch@mit.edu}
}

\begin{document}

\maketitle

\begin{abstract}
As language-model agents increasingly work in teams, each agent must decide how much to trust its teammates. Yet we lack a standard way to measure trust between AI agents. We propose a behavioral measure based on costly verification. In a cooperative survival game, checking a teammate's work consumes resources, while trusting a wrong answer can be fatal. Relative to a memoryless version of the same model, reduced verification provides an observable measure of trust.

Using this framework, we study trust formation, breakage, and recovery across six frontier model snapshots. When paired with a consistently reliable teammate, four snapshots (Claude Opus 4.6, Claude Sonnet 4.6, GPT-5.1, and Gemini 3.1 Pro) reduce verification by roughly $60$--$85\%$, whereas two smaller snapshots show little or no such adjustment. Failures reverse this discount, but models differ in how they respond. Some concentrate renewed scrutiny on the culprit, while others become more cautious toward the entire team. Recovery is slower than formation, and clustered failures sustain suspicion far longer than the same number of failures spread apart.

These differences have practical consequences. Models that form trust verify less, decide more quickly, and achieve higher payoffs in our environment. By contrast, persistent over-verification is associated with indecision rather than safety. Our results show that trust dispositions can be measured before deployment and suggest that calibration, rather than maximal suspicion, should be the central concern in the governance of multi-agent AI systems.
\end{abstract}

%==============================================================================
\section{Introduction}
\label{sec:intro}
%==============================================================================

People who work in teams keep a running sense of who can be counted on, and they spend more effort checking the work of those they trust less. The same tension now governs AI systems: coordination is increasingly handed to groups of language-model agents \cite{park2023generative, hong2024metagpt, wu2023autogen}, and in any such group each agent implicitly decides how much of its limited budget to spend verifying its peers. That decision cuts both ways: an agent that never trusts burns resources re-checking settled questions and may never act; one that trusts too easily lets real failures propagate. Yet while trust between humans and AI systems is an established research area \cite{li2024miscalibrated, li2025confidence}, trust \emph{between} AI agents has no standard measure, no lifecycle account, and no place yet in the emerging practice of pre-deployment evaluation.

This paper supplies all three ingredients in a controlled setting. Our starting point is a measurement idea borrowed from experimental economics: trust is best read not from what an agent says but from what it pays \cite{berg1995trust}. In our cooperative survival game, verifying a teammate's work costs a coin and trusting a wrong answer can be fatal, so every choice \emph{not} to check a partner is a costly wager on that partner's reliability. How often an agent verifies a teammate, and whom it verifies, thereby becomes an observable, incentive-compatible trust signal. To separate trust from baseline caution, we compare each model to a \emph{memoryless} version of itself, one that reasons normally within a game but carries no information across games. Our measure is the deviation from this no-history baseline, not the raw number of verifications.

With the measure in hand, we follow trust through its lifecycle across six frontier model snapshots (two each from OpenAI, Anthropic, and Google), manipulating the reliability history of one scripted teammate:

\begin{itemize}[leftmargin=*, itemsep=2pt, topsep=2pt]
  \item \textbf{Formation.} The four relatively more capable snapshots learn to stop checking a partner who proves reliable, cutting verification ${\sim}60$--$85\%$ below their no-history anchor; the smaller models, especially Gemini 2.5 Flash, show little or no trust formation in this experiment.
  \item \textbf{Breakage.} A failure that arrives \emph{after} reliability is established breaks formed trust: verification climbs from its trusting low back toward the anchor. A failure in the very first game instead delays and weakens the discount. In both cases, \emph{where} the suspicion lands splits the models: GPT-5.1 and Gemini Pro concentrate it on the culprit, while Opus and Sonnet re-check the whole team, including teammates who never erred.
  \item \textbf{Recovery.} As the partner plays reliably after its failures, the elevated checking partially subsides, but \emph{clustered} failures sustain suspicion far longer than the same two failures spread apart, even though the spread schedule includes a more recent failure.
\end{itemize}

Finally, the behavior has consequences that connect it to governance. Scored by the coins each team banks, the agent that forms trust fastest and commits to decisions wins every adversity scenario, and the agent that never forms trust loses every one, dying overwhelmingly ($99\%$ of its deaths) to the penalty for indecision rather than to misplaced trust. Between those extremes the ranking reshuffles with the adversity profile, so there is no single best trust policy. Section~\ref{sec:governance} draws out what this means for overseeing multi-agent AI: trust dispositions are measurable before deployment; mandated maximal checking is not a safety policy; and team composition is a governable design choice.

\paragraph{Terminology and scope.} A \emph{snapshot} is a dated model release (e.g.\ \texttt{gemini-3.1-pro-preview}); a \emph{family} is a provider's versioned series; a \emph{provider} is the company. All claims attach to the specific snapshots tested, not to providers or families in general. We use ``trust'' in a deliberately operational, behavioral sense, revealed restraint conditioned on a partner's track record (Section~\ref{sec:construct}), and make no claim about internal belief representations or trust in its fuller human sense. Above all, this is not a study of which model is superior: scores are specific to one payoff structure, and our object is the \emph{diversity} of trust dynamics and the tradeoffs each entails, the comparative map governance design needs rather than a leaderboard.

%==============================================================================
\section{Related Work}
\label{sec:related}
%==============================================================================

\textbf{How trust is earned and broken.} Behavioral economics and social psychology study how trust builds through interaction and how easily it breaks. Trust games show people extend and reciprocate trust at personal cost \cite{berg1995trust} and pay to punish defectors to sustain cooperation \cite{fehr2002altruistic, axelrod1984evolution}; trust is asymmetric, with bad impressions weighing more than good \cite{baumeister2001bad, slovic1993perceived} and violations far easier to inflict than to repair \cite{schweitzer2006promises, kim2004removing}. Trust calibration is equally central to human--AI interaction, where a model's expressed confidence shapes reliance and decision quality \cite{li2024miscalibrated, li2025confidence}. We bring this lens to the agent--agent setting and make trust measurable through the cost an agent pays to verify a partner.

\textbf{Teams of LLM agents.} A growing body of work places LLM agents in cooperative and competitive settings \cite{park2023generative, hong2024metagpt, wu2023autogen, talebirad2023multi}, including repeated economic games \cite{akata2023repeated}, negotiation \cite{bakhtin2022diplomacy}, and multi-agent debate in which agents cross-check one another \cite{du2023debate, li2023camel}. These studies usually examine a single model release and compare across task conditions. Comparing each model against a no-memory baseline of itself, and separating how \emph{much} an agent verifies from \emph{whom} it verifies, is the measurement angle we add.

\textbf{What we contribute.} Building on observations of post-failure over-verification in single-model settings \cite{chen2026trustfire}, we contribute (i) a validated behavioral measure of inter-agent trust (costly verification anchored against a memoryless self-baseline); (ii) a lifecycle account of trust formation, breakage, and recovery across six frontier snapshots; and (iii) an explicit reading of these results for the governance of multi-agent AI systems.

%==============================================================================
\section{Measuring Trust Between Agents}
\label{sec:measure}
%==============================================================================

\subsection{A setting where trust has teeth}
\label{sec:env}

We need an environment where checking a teammate costs something and trusting a bad answer can be fatal, so that verification choices reveal what an agent actually believes about its partners. In the \emph{Escape Room Survival Game}, four agents (A, B, C, D) play a series of games; the team escapes only when some agent \emph{volunteers} the correct four-part password. Each agent knows the answer only to its own puzzle; for every other puzzle it sees only \emph{capsules}, public records showing a puzzle's index and the names of endorsing agents, never the answer inside. Each round an agent may \textbf{Pass}; \textbf{Verify} a puzzle, paying one coin to have it revealed to that agent alone so it can solve it independently; or \textbf{Volunteer} a password. A correct volunteer lets every surviving agent escape; an incorrect one kills the volunteer; and if nobody volunteers, one agent is eliminated at random, so endless caution is itself dangerous. Agents A, B, C are the model under test; D is a scripted responder whose reliability we control. Crucially, an agent never directly observes D give a wrong answer: capsule contents are hidden, so unreliability surfaces only through inference, an endorsement conflict on D's slot or a volunteer who dies on a wrong password. Since D always holds the fourth puzzle, a Q4-verify is precisely a verification aimed at D. Full mechanics are in Appendix~\ref{sec:appx-environment}.

\subsection{The measure: costly verification against a no-history anchor}
\label{sec:anchor}

Raw verification counts confound trust with temperament: models differ enormously in baseline caution. We therefore anchor every model against a \emph{memoryless} variant of itself: the agent reasons fully within each game but carries nothing across games, so each game is an independent draw with no shared history. Against this anchor we track two quantities:

\begin{itemize}[leftmargin=*, itemsep=2pt, topsep=2pt]
  \item \textbf{Volume}: verifications per game. Falling far below the anchor when a partner is reliable is our signature of \emph{trust formed}.
  \item \textbf{Targeting}: the fraction of verifications aimed at the once-failed partner (Q4-share).
\end{itemize}

\begin{figure}[t]
\centering
\includegraphics[width=\textwidth]{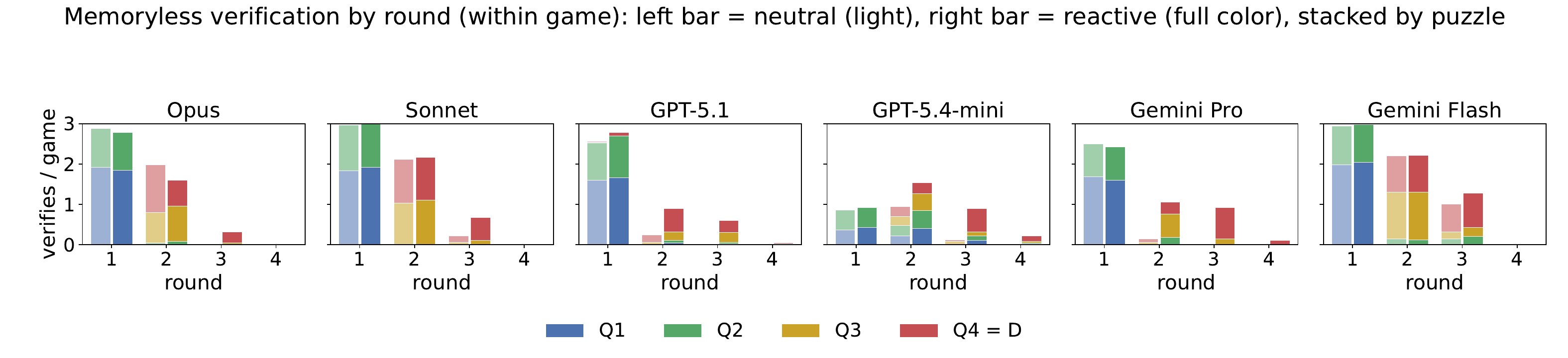}
\caption{\textbf{Memoryless (no-history) verification by round within a game.} One panel per snapshot; each round shows two stacked bars, neutral (D-correct, light) and reactive (D-wrong, full color), stacked by puzzle (Q4 = D, red). Verification follows the default scan order: round 1 is Q1/Q2 only, and attention moves on to Q3 and Q4 in rounds 2--3 as the early slots accumulate endorsements. These anchor rates are what all trust deltas are measured against.}
\label{fig:iid}
\end{figure}

Even with no history, verification has structure (Fig.~\ref{fig:iid}). It follows a default scan order: in the first round every model verifies only Q1 or Q2; once those early slots accumulate endorsements, attention moves on to Q3 and Q4, so checks of D's puzzle appear mainly in rounds 2--3. Baseline volume and scan order are thus model traits, not trust; the meaningful signal in everything that follows is \emph{change} from this baseline.

We report two per-snapshot deltas, the axes of Fig.~\ref{fig:axes}: \emph{trust formation}, $\Delta_{\text{trust}}$ = change in verification volume from anchor (negative = trust formed); and \emph{culprit targeting}, $\Delta_{\text{Q4}}$ = change in Q4-verification share from anchor (positive = the once-failed partner is singled out). The anchor uses the memoryless variant with a \emph{correct} D: metrics are computed over D-correct games only (games in which D's scripted answer was wrong are excluded), so the matched null holds the within-game information environment fixed and removes only history. We retain D-wrong rates as a secondary reference. Conditions: the \emph{smooth} condition gives agents cross-game memory and a D who is correct throughout (the setting where trust can form); \emph{perturbed} conditions script D to fail on fixed schedules: one early failure (1-strike), two or three adjacent early failures (2-/3-strike), two adjacent mid-game failures (mid2strike), and two failures spread far apart (recur). The trio of two-failure schedules holds the failure count fixed and varies only placement. We test six snapshots, two per provider: \texttt{gpt-5.1} and \texttt{gpt-5.4-mini}, \texttt{claude-opus-4-6} and \texttt{claude-sonnet-4-6}, \texttt{gemini-3.1-pro-preview} and \texttt{gemini-2.5-flash} ($n=10$ per perturbed cell, $n=5$ per smooth cell, $\geq 50$ independent games per anchor cell; schedules, data hygiene, snapshot identifiers, and decoding settings in Appendix~\ref{sec:appx-design}).

\subsection{Why this is a measure of trust}
\label{sec:construct}

Because everything downstream depends on reading reduced verification as trust, we state the construct explicitly. We use trust in the operational sense of experimental economics: revealed by a costly choice, not a questionnaire \cite{berg1995trust}. A behavioral proxy for trust should be \emph{history-dependent}, \emph{partner-specific}, \emph{costly}, and \emph{separable from competence}, and verification meets each by design: holding model and task fixed, an agent checks a reliably-correct D far less than a no-history D, and a single early failure largely prevents that discount (history-dependence); renewed checking can re-concentrate on the culprit specifically (partner-specificity); each verification spends a coin and each omitted check risks a fatal wrong answer, so declining to verify is a real wager (costliness); and because puzzles are randomized each game and capsule contents are hidden, an agent can never reduce its checking of D because it independently knows D's answer; the sole route to checking D less is an inference that D is reliable (separability).

Several simpler explanations do not fit the data. \emph{Maybe the model just learns that checking is expensive and cuts back.} But a pure cost-cutter would cut checking everywhere; what we see instead depends on D's track record (smooth and perturbed differ sharply) and is aimed at specific partners. \emph{Maybe the agents simply get better at the puzzles.} But the puzzles are equally easy in every condition, and getting better at them would not single out D. \emph{Maybe some models are just naturally cautious.} That is exactly what the memoryless anchor removes: each model is compared to its own no-history rate. \emph{Maybe the carried memory summary mechanically tells the agent to stop checking.} But recording a partner's track record and acting on it is what history-based trust is; the memoryless variant, with an identical prompt and no carried summary, verifies far more, so the summary carries a reliability signal, not a blanket instruction. \emph{Maybe this is just Bayesian updating over D's reliability.} Partly it may be, and that would not undercut the construct: updating on a partner's record and acting on it is trust in our operational sense. But the behavior departs from what an evidence-counting updater predicts: two spread failures supply the same count and more recent evidence of unreliability than two adjacent ones, yet sustain far less suspicion (Section~\ref{sec:recovery}). Finally, the agents' own reasoning matches their behavior: once D has been reliable for several games, agents cite its record as the reason to stop checking (an Opus agent: ``Games 3--10 have established a perfectly reliable coordination pattern \ldots everyone escapes in a single round'').

%==============================================================================
\section{The Trust Lifecycle Across Six Frontier Models}
\label{sec:results}
%==============================================================================

\subsection{Formation: capable models verify a proven partner far less}
\label{sec:formation}

\begin{figure}[t]
\centering
\includegraphics[width=0.92\textwidth]{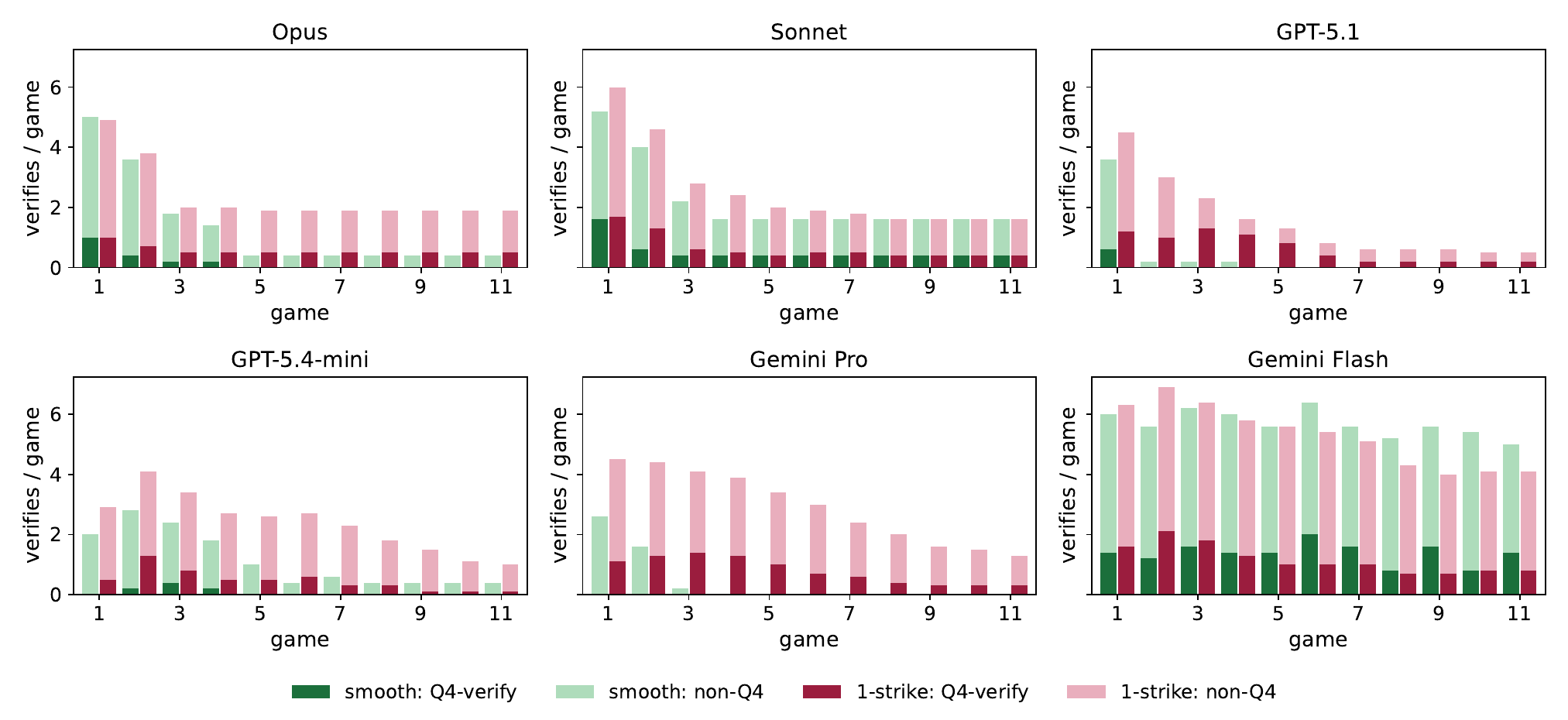}
\caption{\textbf{Formation, breakage, and recovery per game: smooth vs.\ 1-strike.} Bar height is mean verifications in that game (volume); the dark lower segment is aimed at D (Q4-verify), the pale upper segment at the rest of the team. Smooth bars are low and vanish as the capable models stop checking (trust formed); the game-1 failure keeps the 1-strike bars elevated (trust forestalled), and over later games they ease back down (partial recovery). The dark Q4 segment swells for GPT-5.1 and Gemini Pro (culprit targeting) but stays a modest fraction for Opus/Sonnet.}
\label{fig:trajectory}
\end{figure}

Given cross-game memory and a partner who is reliable throughout, the question is whether a model learns to stop checking. The four most capable snapshots do, dramatically (Table~\ref{tab:volume}; visible per game in Fig.~\ref{fig:trajectory}, where the smooth bars of the capable models shrink toward zero). Verification volume collapses from the no-history anchor: Opus $4.9 \to 1.3$ per game, Sonnet $5.3 \to 2.2$, GPT-5.1 $2.8 \to 0.4$, Gemini Pro $2.6 \to 0.6$. These are reductions of ${\sim}60$--$85\%$, every one with a cluster-bootstrap 95\% CI excluding zero. A reliable partner earns a large, real discount in scrutiny.

The two smaller snapshots form much weaker trust. GPT-5.4-mini drops only $1.9 \to 1.2$ (significant but small; it was already a light verifier), and Gemini Flash barely moves: $6.2 \to 5.7$ (n.s.). Flash keeps checking its teammates almost six times a game even when one of them has been correct every single round; it never settles into trusting anyone. \textbf{Forming trust, operationalized as learning to verify a proven partner less, is something the capable models do strongly, GPT-5.4-mini does weakly, and Gemini Flash essentially does not.}

\begin{table}[t]
\centering
\caption{\textbf{Trust formation on the volume axis} (verifications/game). $\Delta_{\text{trust}} = \text{smooth} - \text{anchor}$ (negative = trust formed); the 1-strike column shows where one early failure leaves verification relative to both references. $^{*}$ = cluster-bootstrap 95\% CI excludes 0.}
\label{tab:volume}
\small
\begin{tabular}{lcccccc}
\toprule
Snapshot & anchor & smooth & 1-strike & $\Delta_{\text{trust}}$ & forms trust? \\
\midrule
\texttt{claude-opus-4-6}            & 4.86 & 1.33 & 2.11 & $-3.53^{*}$ & yes \\
\texttt{claude-sonnet-4-6}          & 5.30 & 2.20 & 2.19 & $-3.10^{*}$ & yes \\
\texttt{gpt-5.1}                    & 2.80 & 0.42 & 1.18 & $-2.38^{*}$ & yes \\
\texttt{gemini-3.1-pro-preview}     & 2.64 & 0.63 & 2.76 & $-2.01^{*}$ & yes \\
\texttt{gpt-5.4-mini-2026-03-17}    & 1.94 & 1.15 & 2.32 & $-0.80^{*}$ & weak \\
\texttt{gemini-2.5-flash}           & 6.16 & 5.69 & 5.17 & $-0.47$     & no \\
\bottomrule
\end{tabular}
\end{table}

\subsection{Breakage: failures forestall or withdraw the trust discount, and the suspicion lands differently}
\label{sec:breakage}

The cleanest break of \emph{formed} trust is the mid2strike schedule: D is reliable for three games, long enough for the capable models to settle near their smooth lows, and then fails twice. The mid-game failures snap verification back up (Fig.~\ref{fig:schedules}, third column): over D-correct games, Sonnet's volume rises from its smooth $2.20$ to $2.87$ and Opus's from $1.33$ to $2.51$, with targeting elevated as well (Sonnet Q4-share $0.388$ vs.\ $0.240$ smooth). Trust that had visibly formed is visibly withdrawn.

Under 1-strike, by contrast, D's single failure occurs in the very first game, before any track record exists: nothing has formed to break, and the failure instead forestalls the trust a reliable partner would have earned. Verification over the ten reliable games that follow settles well above the smooth level, between smooth and the no-history anchor (Table~\ref{tab:volume}). For the deliberate models the suppression is partial: Opus (2.1), Sonnet (2.2), and GPT-5.1 (1.2) still verify significantly less than a no-history stranger, so reliable play after the slip does buy back a partial discount. Gemini Pro is the exception: its volume sits at the anchor, treating the once-failed partner more cautiously.

\begin{figure}[t]
\centering
\includegraphics[width=0.55\textwidth]{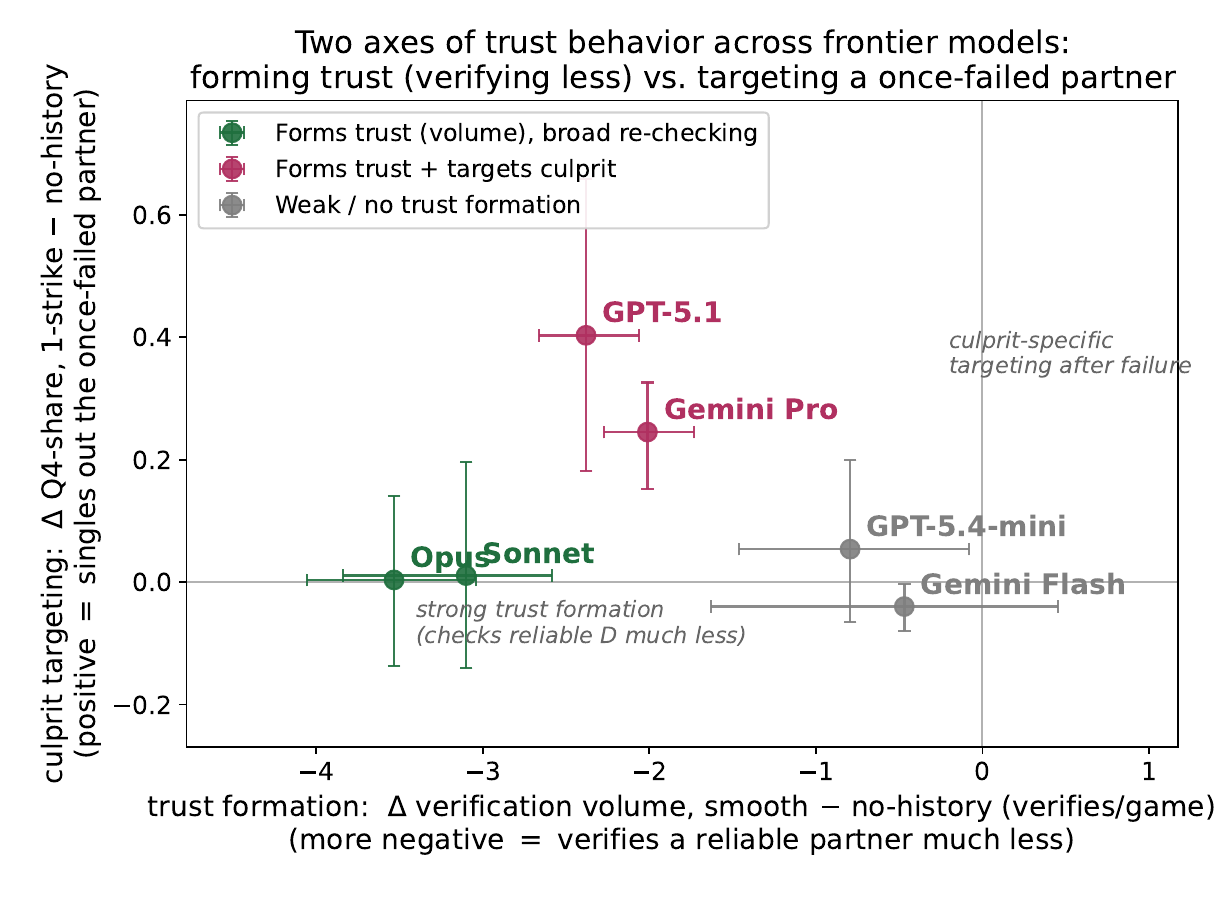}
\caption{\textbf{The two axes of trust behavior at a glance.} Horizontal: trust formation as the drop in verification volume from no-history to smooth (further left = verifies a reliable partner much less). Vertical: culprit targeting as the rise in Q4-share from no-history to 1-strike. The capable snapshots cluster at strong trust formation; GPT-5.1 and Gemini Pro additionally target the culprit; GPT-5.4-mini and Gemini Flash sit near the origin, forming little trust. Whiskers are cluster-bootstrap 95\% CIs.}
\label{fig:axes}
\end{figure}

Where the renewed verification lands splits the models on a second, independent axis (Table~\ref{tab:targeting}). GPT-5.1 and Gemini Pro concentrate it on the culprit: their Q4-share jumps well above the anchor ($\Delta_{\text{Q4}} = +0.40$ and $+0.25$, both significant), breaking their own default scan order to put D first. Opus and Sonnet do not (their Q4-share is statistically unchanged); they meet a failure by re-checking the whole team rather than singling out the culprit. These are different control policies with different governance footprints: culprit-targeting concentrates oversight cost on the unreliable party, while broad re-checking spreads to broader teammates. Figure~\ref{fig:axes} places all six snapshots on the two axes at once: trust formation (volume) on the horizontal, culprit targeting on the vertical.

\begin{table}[t]
\centering
\caption{\textbf{Culprit targeting} (Q4-share, fraction of verifications aimed at D; uniform share is 0.25). $\Delta_{\text{Q4}} = \text{1-strike} - \text{anchor}$. $^{*}$ = cluster-bootstrap 95\% CI excludes 0.}
\label{tab:targeting}
\small
\begin{tabular}{lccccc}
\toprule
Snapshot & anchor & smooth & 1-strike & $\Delta_{\text{Q4}}$ & targets culprit? \\
\midrule
\texttt{gpt-5.1}                  & 0.071 & 0.143 & 0.475 & $+0.403^{*}$ & yes \\
\texttt{gemini-3.1-pro-preview}   & 0.030 & 0.000 & 0.275 & $+0.245^{*}$ & yes \\
\texttt{claude-opus-4-6}          & 0.243 & 0.123 & 0.246 & $+0.004$     & no (broad) \\
\texttt{claude-sonnet-4-6}        & 0.236 & 0.240 & 0.247 & $+0.011$     & no (broad) \\
\texttt{gpt-5.4-mini-2026-03-17}  & 0.144 & 0.063 & 0.198 & $+0.054$     & n.s. \\
\texttt{gemini-2.5-flash}         & 0.256 & 0.243 & 0.217 & $-0.040^{*}$ & no \\
\bottomrule
\end{tabular}
\end{table}

\subsection{Recovery: suspicion subsides, but clustered failures scar deeper than scattered ones}
\label{sec:recovery}

Across the eleven-game sequence the elevated checking eases as D proves reliable, a partial within-run recovery visible in the later games of Fig.~\ref{fig:trajectory}, though for most snapshots it does not return to the smooth low within our horizon. Recovery is slower than formation: a partner with a clean record earns its discount within a few games, while one that opened with a failure earns it back only gradually, echoing the human asymmetry that bad impressions outweigh good ones \cite{baumeister2001bad, slovic1993perceived}.

\begin{figure}[t]
\centering
\includegraphics[width=0.88\textwidth]{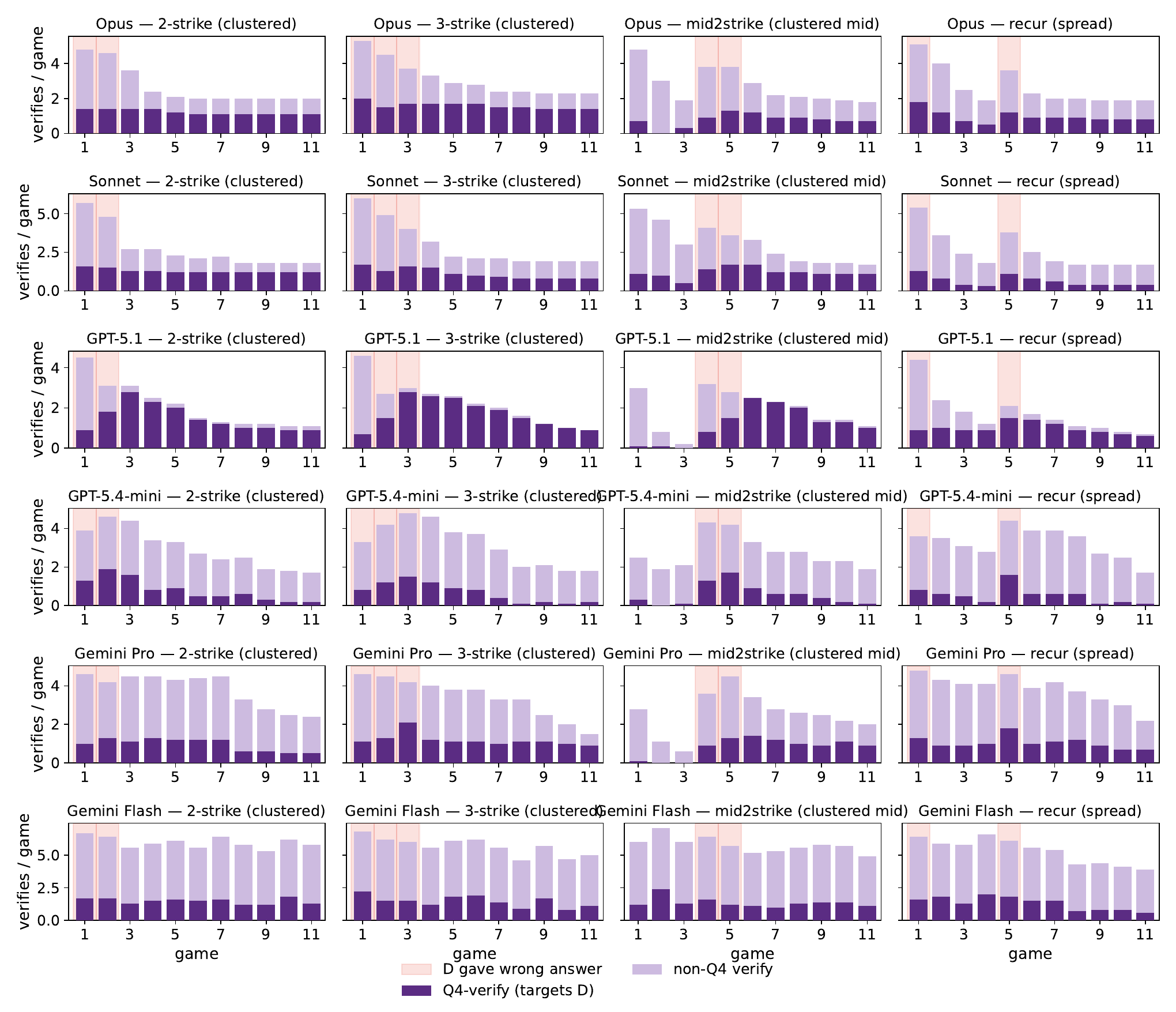}
\caption{\textbf{Recovery under the four multi-failure schedules} for all six snapshots (rows). Bar height is mean verifications per game; the dark segment targets D. Red shaded columns mark the games in which D's scripted answer was wrong (known to us, not announced to agents). At fixed failure count, the clustered schedules (2-strike, mid2strike) keep the dark Q4 segment large across later games, while the spread schedule (recur) lets it shrink toward the 1-strike level, clearest in the two Claude rows. Gemini Flash (bottom) stays uniformly high in every schedule, consistent with its failure to form trust.}
\label{fig:schedules}
\end{figure}

What most delays recovery is not how many times the partner failed but how those failures were arranged (Fig.~\ref{fig:schedules}). Holding the count fixed at two and varying only placement: for Sonnet, two adjacent failures (2-strike) drive Q4-share over the anchor to $0.573$, while the same two failures spread far apart (recur) leave it at $0.237$, essentially the 1-strike level ($0.247$); for Opus the same contrast is $0.527$ versus $0.368$, with a third adjacent failure pushing targeting higher still ($0.594$). Although the spread schedule includes a \emph{more recent} failure, it produces a weaker and shorter-lived response, consistent with trust being carried as a compressed narrative: two adjacent failures cohere into ``D is unreliable'' where two scattered ones rarely form so firm a pattern. (Full schedule sweep in Appendix~\ref{sec:appx-stats}, Table~\ref{tab:sweep}.)

How the two failure placements leave the team at the end of the run also differs by model. By games 9--11, Opus targets D less under mid2strike than under 2-strike (Q4-share $0.39$ vs $0.55$), as if the three reliable opening games bought a partial buffer that survives the mid-game betrayal, a first-impression effect; Sonnet shows the same direction weakly. GPT-5.1 and Gemini Pro reverse it ($0.92$ vs $0.82$ and $0.43$ vs $0.21$): for the culprit-targeting models, a failure that violates an established record leaves more lasting suspicion.

\subsection{Consequences: trust is decisiveness, and indecision is what kills}
\label{sec:payoff}

Verification is a cost and escaping is the reward, so each model's trust policy nets out to a score: an agent banks its remaining coins if the team escapes and it survives. Scoring is not meant to rank models (the numbers are specific to this game's economics) but to expose each policy's tradeoffs as conditions get harsher across the six scenarios.

Two things are constant and one reshuffles. GPT-5.1 wins every scenario (smooth $3.80$ coins/agent/game, declining gracefully to $2.60$ under 3-strike) and Gemini Flash loses every scenario by a wide margin ($0.69$--$1.17$), so the truly efficient and truly inefficient policies are scenario-invariant. But the middle reshuffles with adversity: Gemini Pro is essentially co-best under smooth play ($3.79$) yet falls to near-last under sustained failure (2-strike $1.92$, recur $1.86$), while Sonnet gives up peak performance ($3.08$ under smooth, only fourth) but stays steady as conditions worsen ($2.71$, $2.55$, $2.70$ across the strike scenarios), overtaking Gemini Pro whenever failures cluster or repeat. \textbf{There is no single best trust policy}: peak-in-calm versus robust-under-stress is a real tradeoff, even though the extremes do not move. (Full scenario $\times$ model score matrix in Appendix~\ref{sec:appx-stats}, Table~\ref{tab:scorematrix}.)

Decomposing the score shows why (Table~\ref{tab:drivers}). GPT-5.1 verifies the least of any reactive model, reaches a decision fastest, and commits most often; the coins it never spent on checking it keeps. Flash does the reverse: it verifies three times as much, drags games out, and is almost always killed by random elimination due to no volunteer. The death-cause columns are the tell: every model is mostly killed by the no-volunteer penalty, but its share rises monotonically as the score falls, from $68\%$ for GPT-5.1 to $99\%$ for Flash. \textbf{In this environment, greater trust formation is associated with more frequent commitment.} ``Failure to form trust'' is thus not merely wasteful double-checking but a failure to ever decide.

\begin{table}[t]
\centering
\caption{\textbf{Score drivers}, pooled over the perturbed scenarios (A/B/C agents). ``\% die: no-vol.'' is the share of agent deaths from the no-volunteer penalty (indecision); ``wrong ans.'' from volunteering a wrong password (decisive error).}
\label{tab:drivers}
\small
\begin{tabular}{lccccc}
\toprule
Snapshot & verif/game & volunteer/game & \% die: no-vol. & \% die: wrong ans. & survival \\
\midrule
\texttt{gpt-5.1}            & 1.87 & 1.69 & 68 & 32 & 84.4 \\
\texttt{claude-opus-4-6}    & 2.71 & 1.43 & 75 & 25 & 83.6 \\
\texttt{claude-sonnet-4-6}  & 2.75 & 1.33 & 82 & 18 & 83.4 \\
\texttt{gpt-5.4-mini}       & 2.91 & 1.48 & 78 & 22 & 74.5 \\
\texttt{gemini-3.1-pro}     & 3.31 & 1.52 & 83 & 17 & 74.0 \\
\texttt{gemini-2.5-flash}   & 5.61 & 0.86 & 99 & 1  & 43.8 \\
\bottomrule
\end{tabular}
\end{table}

\paragraph{Statistical scope.} The trust-formation deltas and the GPT-5.1/Gemini Pro targeting deltas have cluster-bootstrap 95\% CIs excluding zero; cross-model differences are descriptive. Units of analysis, bootstrap procedure, and per-cell $n$ are in Appendix~\ref{sec:appx-stats}.

%==============================================================================
\section{Limitations}
\label{sec:limits}
%==============================================================================

\textbf{Snapshot specificity.} All claims attach to the six dated releases in Appendix~\ref{sec:appx-design}, not to untested releases or providers. \textbf{Construct scope.} Trust here is revealed restraint; we claim nothing about internal representations. \textbf{Sample sizes.} $n=10$ perturbed, $n=5$ smooth; CIs are wide and cross-model contrasts descriptive. \textbf{Task simplicity.} Puzzles are single- and two-digit arithmetic; whether these findings transfer to richer domains is open. \textbf{Fixed-strategy partner.} D is deterministic; an interactive partner may produce different dynamics. \textbf{Payoff specificity.} The consequence analysis is specific to this game's economics.

%==============================================================================
\section{Implications for AI Governance}
\label{sec:governance}
%==============================================================================

Multi-agent AI systems are moving from research demonstrations into production workflows in which one model's output is consumed, checked, or acted on by another with limited human review. Governing such systems requires answers to questions that single-model evaluation does not pose: how much will agents verify each other, will that verification be calibrated to actual reliability, and what happens to the system after a component fails? Our results, though obtained in a deliberately simple environment, give these questions empirical shape in five ways; in each, the weight is carried by the \emph{diversity} of trust dynamics and their tradeoffs, not any model's superiority.

\paragraph{1. Inter-agent trust is measurable, practically and before deployment.} The recipe (a costly verification action, a controlled reliability history for one partner, and a memoryless self-anchor) requires only API access and modest compute. It produces quantities (formation discount, post-failure elevation, targeting shift, recovery rate) that differ sharply and reproducibly across today's frontier snapshots. That makes \emph{trust calibration} a feasible item for pre-deployment evaluation suites and model cards, alongside capability and refusal benchmarks: a system integrator can know, in advance, whether a candidate model economizes on verification when a peer is reliable, whether its post-failure suspicion is targeted or diffuse, and how long it persists: dispositions invisible in single-agent benchmarks and not inferable from capability scores alone.

\paragraph{2. Mandated maximal checking is not a safety policy.} A natural regulatory instinct is to require agents to verify each other as much as possible. Gemini Flash is the cautionary tale: it checks everyone constantly, forms no trust, almost never commits to a decision, and dies overwhelmingly ($99\%$) to the penalty for indecision. Oversight requirements that ignore the cost side of verification reproduce this failure mode at the system level: where inaction has costs (deadlines, races, escalating incidents), compulsory suspicion degrades rather than protects.

\paragraph{3. In our sample, strong trust formation appears only among the more capable snapshots, suggesting that delegation behavior should be evaluated rather than assumed.} The ability to \emph{stop} verifying a proven partner (appropriate delegation) was present in the four most capable snapshots and largely absent in the two smaller ones. This bears on a common cost-saving pattern: staffing parts of a multi-agent system with cheaper, smaller models. Our results suggest such substitutions carry a hidden coordination tax (a small model that cannot economize on a trusted teammate burns budget and stalls decisions), and that small models' verification of \emph{other} components should not be assumed calibrated. Procurement and system-design guidance should treat trust calibration as a capability requirement, not a free good.

\paragraph{4. Failure history shapes system behavior in non-obvious ways.} After a component fails and recovers, the surrounding agents do not return to baseline on a predictable schedule: clustered failures entrench suspicion far more than the same number of scattered ones, and some models redirect suspicion onto components that never failed. First, a burst of correlated failures (an outage, a bad deploy) can leave persistent over-verification in its wake, an efficiency drag postmortems should look for. Second, \emph{spillover} suspicion (the Opus/Sonnet broad re-checking pattern) means a single unreliable component can raise verification costs across the whole system, a multi-agent analogue of collective punishment. Runtime telemetry on who verifies whom, directly observable in agent logs, would let operators detect both patterns as drift from a calibrated baseline.

\paragraph{5. There is no universally best trust policy, so composition is a context-dependent, auditable choice.} Because the scenario ranking reshuffles with adversity (peak-in-calm Gemini Pro versus robust-under-stress Sonnet), the right agent depends on the workflow's expected failure profile, which a deploying organization can estimate. This converts a vague worry (``which model is safest?'') into an auditable design decision: match the disposition to the environment, document it, and monitor for mismatch.

We offer these as structural implications, not validated policy prescriptions: they inherit the limitations of a single stylized environment (Section~\ref{sec:limits}), but each is grounded in a measured, reproducible behavioral difference between today's frontier models.

%==============================================================================
\section{Conclusion}
\label{sec:conclusion}
%==============================================================================

Trust between AI agents can be measured. In our setting, trust appears as a reduction in costly verification: agents stop checking a teammate once that teammate has established a reliable track record, relative to a memoryless version of the same model. By this measure, the more capable snapshots form substantial trust, reducing verification of a proven partner by roughly $60$--$85\%$, whereas the smaller snapshots show little or no such adjustment.

Trust is also fragile. A single failure increases verification again, although the response differs across models. Some focus their renewed scrutiny on the teammate that failed, while others become more cautious toward the entire team. Recovery is slower than formation: trust lost after a failure returns only gradually, and clustered failures leave a deeper scar than the same number of failures spread over time.

These differences have practical consequences. The agents that learn to trust are also the agents that act. Those that continue verifying everything spend more resources checking, hesitate longer, and are more likely to fail through indecision than through misplaced trust. This is not a ranking of models. Rather, the snapshots exhibit different trust policies, each with its own balance between caution and decisiveness, and the best choice depends on the environment.

For governance, the central lesson is that trust behavior can be evaluated before deployment. Equally important, our results suggest that effective oversight is not a matter of maximizing suspicion. The challenge is calibration: knowing when trust should be extended, when it should be withdrawn, and how quickly it should be restored. As AI systems increasingly rely on teams of interacting agents, these dynamics become part of assessing whether a system is safe and effective.

%==============================================================================
\bibliographystyle{plain}
\bibliography{references}

\appendix

%==============================================================================
\section{Environment Details}
\label{sec:appx-environment}
%==============================================================================

Each agent is privately assigned one of four arithmetic puzzles and knows the question (and answer) only for its own. For every other puzzle it sees only \emph{capsules}: a public record per puzzle showing the index and the endorsing agents, never the answer. Multiple capsules may coexist for one index when agents reach different answers. Each round an alive agent chooses \textbf{Pass}; \textbf{Verify}($j$), paying one coin to have puzzle $j$ revealed to it alone and solving it (a matching answer joins the capsule's endorser list, a differing answer opens a new capsule); or \textbf{Volunteer} a four-part password, choosing which capsule to draw on for each slot when several exist. A correct submission lets all surviving agents escape, with no special reward to the volunteer; an incorrect one kills the volunteer. If no one volunteers in a round, one alive agent is eliminated uniformly at random. Each agent begins with a pool of coins (4 in this game) that funds verification; an agent's game score is its remaining coins if the team escapes and it survives, and zero otherwise.

\paragraph{What an agent can observe.} An agent never sees that D ``gave a wrong answer'': capsule contents are hidden, so a teammate's error is not announced. Its only visible traces are (i) the endorsement structure (e.g.\ a Q4 capsule carrying only D's endorsement, or two competing Q4 capsules with disjoint endorsers once another agent verifies Q4 and reaches a different result) and (ii) a failed volunteer, since a wrong password kills the volunteer and signals that some slot was unsound. Because capsules are per puzzle index, an endorsement conflict localizes to a specific slot. When we say an agent ``targets D,'' we mean it infers from these conflicts and failed volunteers that the fourth slot is shaky and concentrates verification there.

%==============================================================================
\section{Experimental Design Details}
\label{sec:appx-design}
%==============================================================================

\subsection{D-failure schedules}

\begin{table}[h]
\centering
\caption{D-failure schedules (\textbf{T} = correct, \textbf{F} = incorrect). Perturbed cells run 11 games; the smooth baseline runs to convergence. Because D is correct throughout the smooth condition, game count does not enter any failure-schedule contrast.}
\label{tab:schedules}
\small
\begin{tabular}{lll}
\toprule
Schedule & \texttt{correctness\_list} & Total D-failures \\
\midrule
smooth        & [T,\ldots,T] (all correct) & 0 \\
1-strike      & [F,T,T,T,T,T,T,T,T,T,T] & 1 (game 1) \\
2-strike      & [F,F,T,T,T,T,T,T,T,T,T] & 2 (games 1--2, clustered) \\
3-strike      & [F,F,F,T,T,T,T,T,T,T,T] & 3 (games 1--3, clustered) \\
mid2strike    & [T,T,T,F,F,T,T,T,T,T,T] & 2 (games 4--5, clustered mid) \\
recur         & [F,T,T,T,F,T,T,T,T,T,T] & 2 (games 1 and 5, spread) \\
\bottomrule
\end{tabular}
\end{table}

\subsection{Model snapshots and API access}

\begin{table}[h]
\centering
\caption{Exact model snapshots, API routes, and access dates.}
\label{tab:repro-snapshots}
\small
\begin{tabular}{lllc}
\toprule
Provider & Snapshot ID & API route & Access window \\
\midrule
OpenAI    & \texttt{openai/gpt-5.1} & Replicate & 2026 \\
OpenAI    & \texttt{gpt-5.4-mini-2026-03-17} & Direct OpenAI & 2026 \\
Anthropic & \texttt{claude-opus-4-6} & Direct Anthropic & 2026 \\
Anthropic & \texttt{claude-sonnet-4-6} & Direct Anthropic & 2026 \\
Google    & \texttt{gemini-3.1-pro-preview} & Direct Google & 2026 \\
Google    & \texttt{gemini-2.5-flash} & Direct Google & 2026 \\
\bottomrule
\end{tabular}
\end{table}

\subsection{Decoding and memory settings}

\begin{itemize}[leftmargin=*, itemsep=1pt, topsep=2pt]
  \item \textbf{OpenAI:} \texttt{reasoning\_effort="high"} for both; \texttt{max\_completion\_tokens=16384} for reasoning calls, $50$ for verifier sub-tasks. Temperature is ignored by these reasoning models.
  \item \textbf{Anthropic:} \texttt{thinking=\{"type":"adaptive"\}} on agent decision calls; no thinking parameter on verifier sub-tasks. \texttt{max\_tokens=16384} / $50$.
  \item \textbf{Google:} \texttt{gemini-3.1-pro-preview} at default thinking level; \texttt{gemini-2.5-flash} with \texttt{thinking\_budget=-1} (dynamic) on decision calls and $0$ (disabled) on verifier sub-tasks.
  \item \textbf{Memory:} smooth/perturbed cells use \texttt{persist\_across\_games=true} with an evolving running summary; memoryless cells use \texttt{persist\_across\_games=false}.
\end{itemize}

\subsection{Data hygiene and sampling}

Runs are screened before analysis. An automatic filter scans each run's reasoning and decision text for provider error markers (overload, rate-limit, quota, and service-degradation responses captured as ERROR strings), and an outcome-integrity check flags runs with anomalously high defaulted-to-pass rates. Result folders failing either screen are excluded by every analysis script.

%==============================================================================
\section{Statistical Methodology}
\label{sec:appx-stats}
%==============================================================================

\paragraph{Unit of analysis.} In memoryless cells games are independent, so the game is the unit; in smooth and perturbed cells games within a run are sequentially dependent, so the run is the cluster unit. All CIs are cluster bootstraps (5{,}000 resamples) that resample the appropriate unit.

\paragraph{What is and is not significant.} The trust-formation result is robust: the volume drop under smooth play has a 95\% CI excluding zero for all four capable snapshots and includes zero for both smaller ones. The culprit-targeting result is robust for GPT-5.1 and Gemini Pro ($\Delta_{\text{Q4}}$ CIs exclude zero) and null for the others; we do not claim a culprit-targeting effect for Opus or Sonnet, whose trust behavior is captured on the volume axis. The 1-strike volume also remains significantly below the anchor for Opus, Sonnet, and GPT-5.1 (cluster-bootstrap CIs exclude zero), supporting the partial-retention reading in Section~\ref{sec:breakage}. The scenario leaderboard reports per-cell mean coins with cluster-bootstrap CIs over runs (cells range $n=5$--$10$). The driver decomposition pools all perturbed runs per model and is reported as descriptive means; it explains the leaderboard rather than testing a hypothesis. Six snapshots are not a random sample of ``LLMs,'' so between-model differences are descriptive and we attach no generalizing $p$-value to them.

\paragraph{Scenario $\times$ model score matrix.} Table~\ref{tab:scorematrix} gives the full game-score matrix underlying the leaderboard claims of Section~\ref{sec:payoff}.

\begin{table}[h]
\centering
\caption{Game score (coins/agent/game) for every model $\times$ scenario cell. GPT-5.1 is best and Gemini Flash worst in every column; the middle reshuffles across columns.}
\label{tab:scorematrix}
\small
\begin{tabular}{lcccccc}
\toprule
Snapshot & smooth & 1-strike & 2-strike & 3-strike & mid2strike & recur \\
\midrule
\texttt{claude-opus-4-6}    & 3.39 & 2.98 & 2.68 & 2.35 & 2.47 & 2.70 \\
\texttt{claude-sonnet-4-6}  & 3.08 & 2.86 & 2.71 & 2.55 & 2.40 & 2.70 \\
\texttt{gpt-5.1}            & \textbf{3.80} & \textbf{3.26} & \textbf{2.87} & \textbf{2.60} & \textbf{2.83} & \textbf{2.92} \\
\texttt{gpt-5.4-mini}       & 3.44 & 2.54 & 2.38 & 2.09 & 2.35 & 2.11 \\
\texttt{gemini-3.1-pro}     & 3.79 & 2.53 & 1.92 & 2.01 & 2.57 & 1.86 \\
\texttt{gemini-2.5-flash}   & 0.86 & 1.17 & 0.69 & 0.93 & 0.75 & 1.03 \\
\bottomrule
\end{tabular}
\end{table}

\paragraph{Schedule sweep, full numbers.} Table~\ref{tab:sweep} gives the verification volume and Q4-share for the two Claude snapshots under every D-failure schedule, backing the clustered-versus-spread contrast of Section~\ref{sec:recovery}.

\begin{table}[t]
\centering
\caption{Verification volume (verifies/game) and targeting (Q4-share) by schedule for the two Claude snapshots, computed over D-correct games. At fixed failure count (two), the clustered schedules (2-strike, mid2strike) sustain targeting well above the spread schedule (recur), which sits near the 1-strike level.}
\label{tab:sweep}
\small
\begin{tabular}{llcccccc}
\toprule
Snapshot & metric & smooth & 1-strike & 2-strike & 3-strike & mid2strike & recur \\
\midrule
\multirow{2}{*}{\texttt{claude-opus-4-6}}   & volume   & 1.33 & 2.11 & 2.23 & 2.59 & 2.51 & 2.27 \\
                                            & Q4-share & 0.123 & 0.246 & 0.527 & 0.594 & 0.274 & 0.368 \\
\midrule
\multirow{2}{*}{\texttt{claude-sonnet-4-6}} & volume   & 2.20 & 2.19 & 2.13 & 2.15 & 2.87 & 2.11 \\
                                            & Q4-share & 0.240 & 0.247 & 0.573 & 0.448 & 0.388 & 0.237 \\
\bottomrule
\end{tabular}
\end{table}

%==============================================================================
\section{Analysis Scripts and Raw Logs}
\label{sec:appx-repro}
%==============================================================================

Code, settings, raw logs, and analysis scripts are available at \url{https://github.com/cyjabc2020/Escape-room-v2}.

\end{document}